\documentclass[conf]{new-aiaa}
\usepackage[utf8]{inputenc}

\usepackage{graphicx}
\usepackage{amsmath}
\usepackage[version=4]{mhchem}
\usepackage{siunitx}
\usepackage{longtable,tabularx}
\title{Motivations and Preliminary Design for\\Mid-Air Deployment of a Science Rotorcraft on Mars}

\author{
Jeff Delaune\footnote{Robotics Technologist, Mobility and Robotic Systems Section, JPL, \href{mailto:jeff.h.delaune@jpl.nasa.gov}{jeff.h.delaune@jpl.nasa.gov}}
Jacob Izraelevitz\footnote{Robotics Technologist, Mobility and Robotic Systems Section, JPL, \href{mailto:jacob.izraelevitz@jpl.nasa.gov}{jacob.izraelevitz@jpl.nasa.gov}}
Evgeniy Sklyanskiy\footnote{Mission Design Engineer, Guidance and Control Section, JPL, \href{mailto:evgeniy.sklyanskiy@jpl.nasa.gov}{evgeniy.sklyanskiy@jpl.nasa.gov}}
Aaron Schutte\footnote{Robotics Systems Engineer, Mobility and Robotic Systems Section, JPL, \href{mailto:aaron.d.schutte@jpl.nasa.gov}{aaron.d.schutte@jpl.nasa.gov}}
Abigail Fraeman\footnote{Scientist, Planetary Sciences Section Section, JPL, \href{mailto:abigail.a.fraeman@jpl.nasa.gov}{abigail.a.fraeman@jpl.nasa.gov}}
Valerie Scott\footnote{Microdevices Engineer, Microdevices Laboratory, JPL, \href{mailto:valerie.j.scott@jpl.nasa.gov}{valerie.j.scott@jpl.nasa.gov}}
Carl Leake\footnote{NASA Space Technology Research Fellow, Mobility and Robotic Systems Section, JPL, \href{mailto:carl.leake@jpl.nasa.gov}{carl.leake@jpl.nasa.gov}}
Erik Ballesteros\footnote{Engineer, JPL, \href{mailto:erik.n.ballesteros@jpl.nasa.gov}{erik.n.ballesteros@jpl.nasa.gov}}
Kim Aaron\footnote{Chief Engineer for Architecture and Formulation, Payload \& Small Spacecraft Mechanical Engineering Section, JPL, \href{mailto:kim.m.aaron@jpl.nasa.gov}{kim.m.aaron@jpl.nasa.gov}}
}
\affil{Jet Propulsion Laboratory, California Institute of Technology, Pasadena, CA, 91109}

\author{
Larry A. Young\footnote{Aerospace Engineer, Aeromechanics Office, \href{mailto:larry.a.young@nasa.gov}{larry.a.young@nasa.gov}, AIAA Associate Fellow}
Wayne Johnson\footnote{Aerospace Engineer, Aeromechanics Office, \href{mailto:wayne.johnson@nasa.gov}{wayne.johnson@nasa.gov}, AIAA Fellow}
Shannah Withrow-Maser\footnote{Engineer, Aeromechanics Office, \href{mailto:shannah.n.withrow@nasa.gov}{shannah.n.withrow@nasa.gov}, AIAA Member}
Haley Cummings\footnote{Engineer, Aeromechanics Office, \href{mailto:haley.cummings@nasa.gov}{haley.cummings@nasa.gov}, AIAA Member}
Raghav Bhagwat\footnote{Intern, Aeromechanics Office, \href{mailto:bhagwat.14@buckeyemail.osu.edu}{bhagwat.14@buckeyemail.osu.edu}}
}
\affil{NASA Ames Research Center, Moffett Field, CA, 94035}

\author{
Marcel Veismann\footnote{Ph.D. student, Department of Aerospace, \href{mailto:mveisman@caltech.edu}{mveisman@caltech.edu}}
Skylar Wei\footnote{Ph.D. Student, Computing + Mathematical Sciences Department, \href{mailto:swei@caltech.edu}{swei@caltech.edu}}
Regina Lee\footnote{Student, Mechanical and Civil Engineering, \href{mailto:relee@caltech.edu}{relee@caltech.edu}}
Luis Pabon Madrid\footnote{Student, Mechanical and Civil Engineering, \href{mailto:lpabonma@caltech.edu}{lpabonma@caltech.edu}}
Morteza Gharib\footnote{Professor \& Director, Aeronautics and Bioinspired Engineering, \href{mailto:mgharib@caltech.edu}{mgharib@caltech.edu}}
Joel Burdick\footnote{Professor, Mechanical and Civil Engineering, \href{mailto:jwb@robotics.caltech.edu}{jwb@robotics.caltech.edu}}
}
\affil{Center for Autonomous Systems and Technologies (CAST), California Institute of Technology, Pasadena, CA, 91125}

\author{William Rapin\footnote{Research Fellow, IRAP, \href{mailto:william.rapin@irap.omp.eu}{william.rapin@irap.omp.eu}}}
\affil{CNRS, Toulouse, 31400, France}

\begin{document}

\maketitle

\begin{abstract}
Mid-Air Deployment (MAD) of a rotorcraft during Entry, Descent and Landing (EDL) on Mars eliminates
the need to carry a propulsion or airbag landing system. This reduces the total mass inside the aeroshell
by more than 100 kg and simplifies the aeroshell architecture. MAD's lighter and simpler design is likely to bring the
risk and cost associated with the mission down. Moreover,
the lighter entry mass enables landing in the Martian highlands, at elevations inaccessible to current EDL
technologies. This paper proposes a novel MAD concept for a Mars helicopter. We suggest a minimum science
payload package to perform relevant science in the highlands. A variant of the Ingenuity helicopter is
proposed to provide increased deceleration during MAD, and enough lift to fly the science payload
in the highlands. We show in simulation that the lighter aeroshell results in a lower terminal velocity (30 m/s)
at the end of the parachute phase of the EDL, and at higher altitudes than other approaches. After discussing
the aerodynamics, controls, guidance, and mechanical challenges associated with deploying at such speed, we propose
a backshell architecture that addresses them to release the helicopter in the safest conditions. Finally, we
implemented the helicopter model and aerodynamic descent perturbations in the JPL Dynamics and Real-Time Simulation (DARTS) framework.
Preliminary performance evaluation indicates landing and helicopter operations can be achieved up to +5 km MOLA (Mars Orbiter Laser Altimeter reference).
\end{abstract}

\section{Nomenclature}

{\renewcommand\arraystretch{1.0}
\noindent\begin{longtable*}{@{}l @{\quad=\quad} l@{}}
$\sigma$ & Rotor solidity \\
$C_T$ & Thrust coefficient \\
$C_Q$ & Torque coefficient \\
$M$ & Mach number \\
$M_{tip}$ & Blade tip Mach number \\
$R$ & Rotor radius \\
$A$ & Rotor disk area \\
$\alpha$ & Angle of attack \\
$v_i$ & Rotor induced velocity \\
$v_h$ & Rotor induced velocity at hover \\
$f$ & Factor that allows the instability in VRS to be reduced or suppressed \\
$k$ & Factor to account for additional induced losses \\
$\Omega$ & Rotor angular velocity \\
$\Omega_{max}$ & Maximum rotor angular velocity \\
$\tau_{max}$ & Maximum rotor torque \\
$\rho$ & Atmospheric density \\
$v_w$ & Wind velocity \\ 
$\theta_{max}$ & Maximum blade collective \\
$n_s$ & stall parameter \\
$\left(\frac{C_T}{\sigma}\right)_s$ & Stall limit thrust coefficient ratio  \\
$C_D$ & Coefficient of drag for the helicopter base \\
$s$ & Side length of the cube-shaped base of the helicopter \\
$\mu$ & Rotor advance ratio \\
$\lambda$ & Rotor inflow ratio \\
$\boldsymbol{V}$ & Flow-relative rotorcraft velocity (in the far field)

\end{longtable*}}

\section{Introduction}

Mid-Air Deployment (MAD) is a novel Entry, Descent and Landing (EDL) technology that enables the transition of a rotorcraft from a stowed configuration
inside the aeroshell during atmospheric descent, to stable controlled flight above the surface of Mars before landing.

The \emph{Ingenuity} Mars helicopter is on its way to attempt the first powered airborne flight on another planet
in 2021~\cite{Balaram2018}. If this rotorcraft can be controlled to safely fly above the surface of Mars, this will pave the way for a new range of aerial robotic
explorers, with regional-scale mobility. An on-going \emph{Mars Science Helicopter} effort at JPL and NASA Ames is investigating various such helicopter designs and
their payload capability~\cite{Withrow2020,Johnson2020}.

Ingenuity will be attached under the \emph{Perseverance} rover until it finds a safe terrain to fly
above. Delivering a helicopter-only efficiently is becoming an increasingly relevant problem to solve,
given the uncertainty surrounding future Mars robotic exploration budget. There is no large landing mission planned
after the \emph{Mars Sample Return} mission~\cite{Muirhead2018}.

For a helicopter-only mission, MAD eliminates the need to carry the propulsion or airbag landing system traditionally used in Mars EDL~\cite{Braun2006}. This saves 100+
kg in entry mass and reduces the complexity of the aeroshell design. This simpler and lighter design is likely to reduce
the mission costs. Additionally, more room in the aeroshell means it can accommodate
a larger rotorcraft with enhanced performance and/or more payload.

The lighter entry mass enables faster deceleration during entry and landing at higher elevations than current EDL
capabilities, potentially above +5 km MOLA (Mars Orbiter Laser Altimeter reference). MAD therefore enables unique and key in situ science in the currently
inaccessible ancient Martian highlands. Rotorcraft can also traverse larger distances (100+ km for the whole mission~\cite{Johnson2020}) and rougher terrains than
rovers. Even with a low-mass payload, these advantages would allow this mission to address high-priority questions
related to the cessation of the Martian dynamo, the planet’s magmatic evolution, the existence of a differentiated
continental crust, the habitability and the nature of weathering/climate during the earliest period of Mars’ history.

However the MAD maneuver must be designed carefully to ensure the helicopter can be released safely away from the
aeroshell, be controllable and avoid dangerous descent aerodynamic conditions.

\subsection{Concept of Operations}
\begin{figure}[hbt!]
\centering
\includegraphics[width=\textwidth]{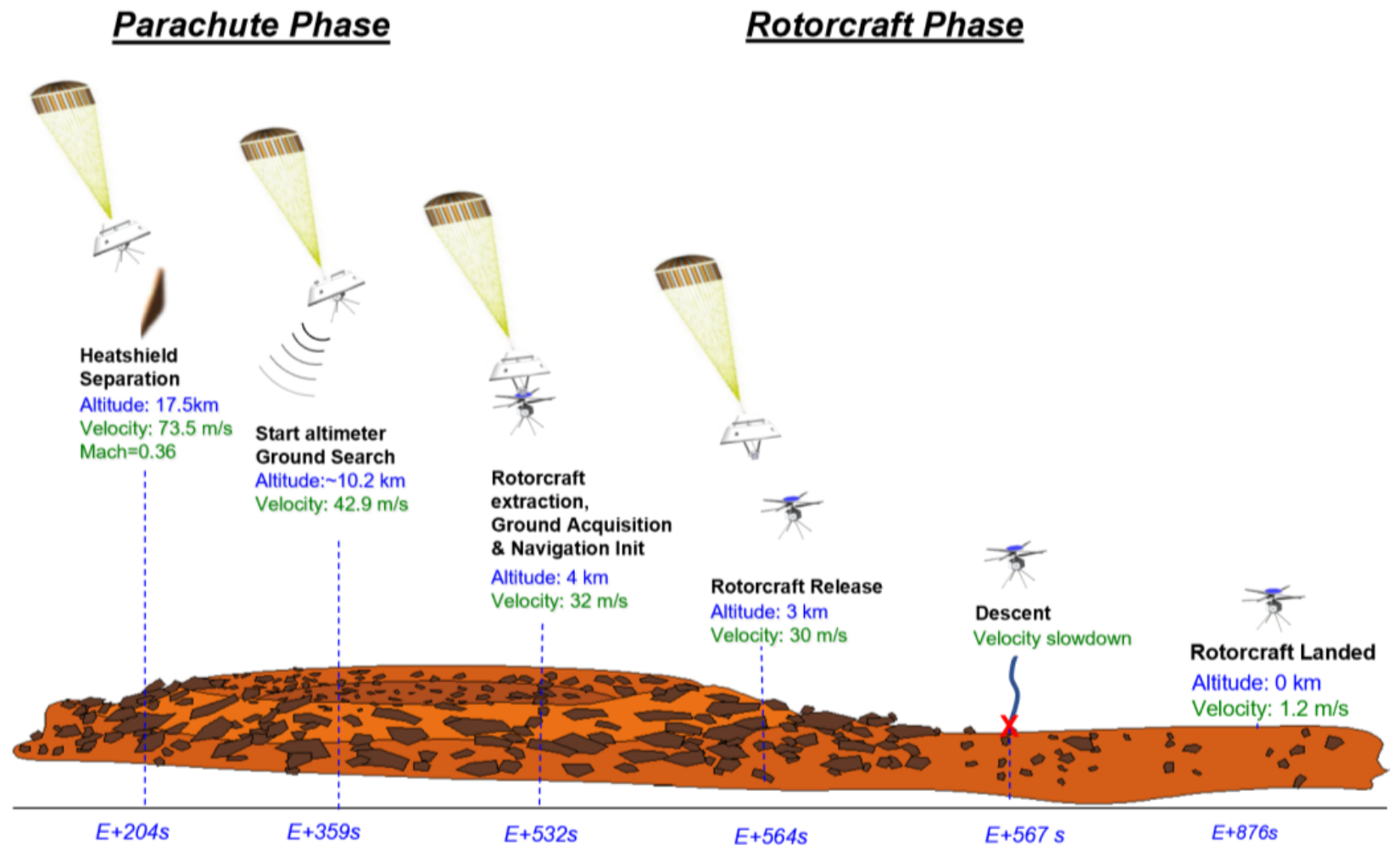}
\caption{Mid-air deployment concept of operations. Velocity and altitude are based on simulation results. Altitudes are expressed with respect to landing elevation.}
\label{fig:conops}
\end{figure}

Our baseline MAD concept, illustrated in Fig.~\ref{fig:conops}, relies on the flight-proven 70-deg sphere-cone aeroshell and parachute
up to heatshield jettisoning~\cite{Braun2006}. Once the \{chute+backshell\} system reaches its terminal velocity, a linkage mechanism extracts the rotorcraft
from the backshell and places it at a pre-determined angle of attack with respect to the airflow. Vision-based navigation is
initialized~\cite{Delaune2020}, and the rotors are spun up while still attached to the backshell. The rotorcraft release is altitude-triggered using
the backshell altimeter. The helicopter descent trajectory is controlled to not only clear away from the backshell, but also
avoid the uncontrollable aerodynamic descent flight regimes. After a safe touchdown, battery recharge and system checks, the
rotorcraft is ready to proceed with nominal mission operations.

Variants of this concept are discussed in our companion paper~\cite{Young2020}, with a focus on the aerodynamic flow conditions.

\subsection{State of the Art}

APL’s Dragonfly is set to attempt the first planetary MAD on Titan in 2034~\cite{Lorenz2018}. Because of the higher descent terminal velocity
(> 30 m/s) and lower atmospheric density (1\% Earth), MAD is significantly more challenging on Mars than on Titan (2 m/s, 4.4 Earth density).
Not only must deployment and helicopter descent occur in a faster timeline on Mars, the rotorcraft will also face challenging aerodynamic
descent perturbations such as Vortex Ring States (VRS)~\cite{JohnsonVRS}. Ingenuity will be the first planetary rotorcraft, but it will be deployed
on the ground after a joint EDL with the Mars 2020 mission. Successful Mars landers and rovers have been soft-landed at elevations up to -1.4 km MOLA for the Opportunity rover
by relying on propulsive or airbags landing systems~\cite{Braun2006}. Similar designs could be used for rotorcraft-only Mars missions , but
these introduce extra complexity, mass, risk, and cost to the mission with respect to MAD. Since MAD has the lowest entry mass, it enables
landing at higher elevations than other options.

Aerially-deployed autonomous aircraft on Earth are at different levels of maturity depending on the type of platform. Fixed-wing aircraft that deploy mid-air are in active use by the US military~\cite{Raytheon}, including deployment from larger aircraft~\cite{NRL}.
Recent investments by DARPA have pushed the development of helicopters that can also be aerially deployed, both coaxial~\cite{Ascent} and
multirotor~\cite{Bouman2020}. However, MAD with a Mars helicopter faces a significantly different control challenge than on Earth~\cite{Grip2018}, further
stressed by the VRS conditions mentioned earlier.

\subsection{Contributions}

This paper summarizes the preliminary findings of an-going effort to investigate mid-air deployed Mars rotorcraft,
between between NASA's Jet Propulsion Laboratory and Ames Research Center. Our contributions are:
\begin{itemize}
	\item a novel EDL concept for MAD that reduces the complexity and mass of the backshell, which could increase the maximum
	    landing elevation on Mars by several kilometers (> 5 km MOLA) based on preliminary results;
	\item relevant science objectives, target locations and payload instrument candidates for a rotorcraft in the Martian Highlands;
	\item a rotorcraft design optimized for flight at low air density, both for high-altitude deployment and highland operations. This
	    design is further described in our companion paper~\cite{Young2020};
	\item entry and descent architecture and simulation results for the MAD scenario, including comparison with past Mars mission scenarios;
	\item a backshell architecture that avoids recontact and allows the rotor to be spun up and tilted prior to release for stable transition to powered flight;
	\item helicopter deployment and descent preliminary simulation results, including aerodynamic vortex ring state descent perturbations.
\end{itemize}

Section~\ref{sec:science} discusses the science motivations and defines a representative instrument payload. Section~\ref{sec:heli} summarizes the
highland rotorcraft design, which is the focus of our companion paper~\cite{Young2020}. Section~\ref{sec:EDL} is focused the EDL simulation results
and comparison. Finally, Section~\ref{sec:mad-design} discusses the MAD challenges and introduces a backshell deployment mechanism to alleviate them,
while Section~\ref{sec:mad-sim} shows the preliminary deployment simulation and experimental results.

\section{Science Motivation: Accessing the Highlands}
\label{sec:science}

\begin{figure}[hbt!]
\centering
\includegraphics[width=\textwidth]{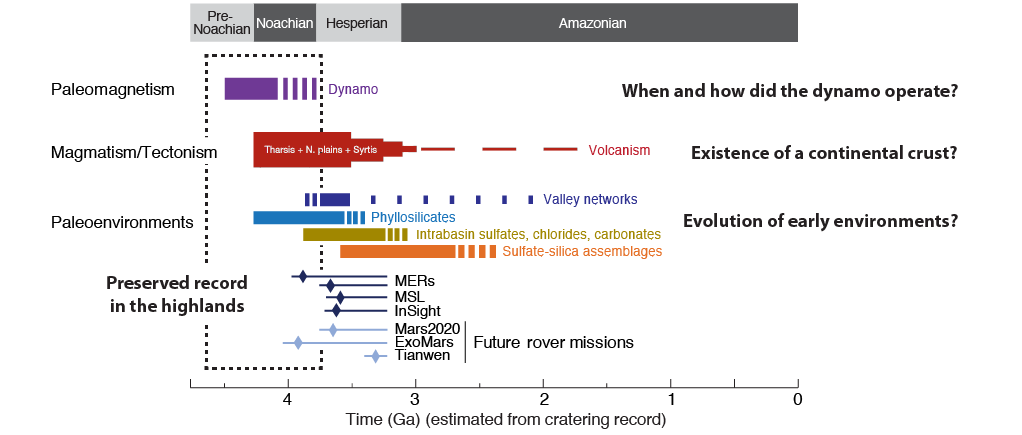}
\caption{Timeline of major processes in Mars history and related outstanding questions (modified after Ehlmann \& Edwards, 2014). Estimated crater age of terrains for landed missions with error bars indicating uncertainty and extended to the right to highlight subsequent surface alteration.}
\label{fig:objectives}
\end{figure}

Mars’ southern highlands are the oldest well-preserved record of major geological evolution on early Mars, and can solve key outstanding questions in planetary science today \cite{Rapin2020} (Fig.~\ref{fig:objectives}): (i) How did the Martian dynamo operate? (ii) Which processes formed the early crust of Mars? (iii) How did early environments evolve? Yet, despite their importance, the oldest terrains have never been explored \emph{in situ} primarily due to current entry descent and landing (EDL) technology limitations.

MAD rotorcraft are a unique technology that will enable measurements relevant to each of these science questions. Specifically, MAD will be transformative over current flagship rover designs because it will enable \emph{in situ} access to the Martian highlands at regional scales (100+ km on a mission lifetime), for reduced cost, bridging the gap between orbital maps and detailed \emph{in situ} analyses.

\begin{figure}[b!]
\centering
\includegraphics[width=\textwidth]{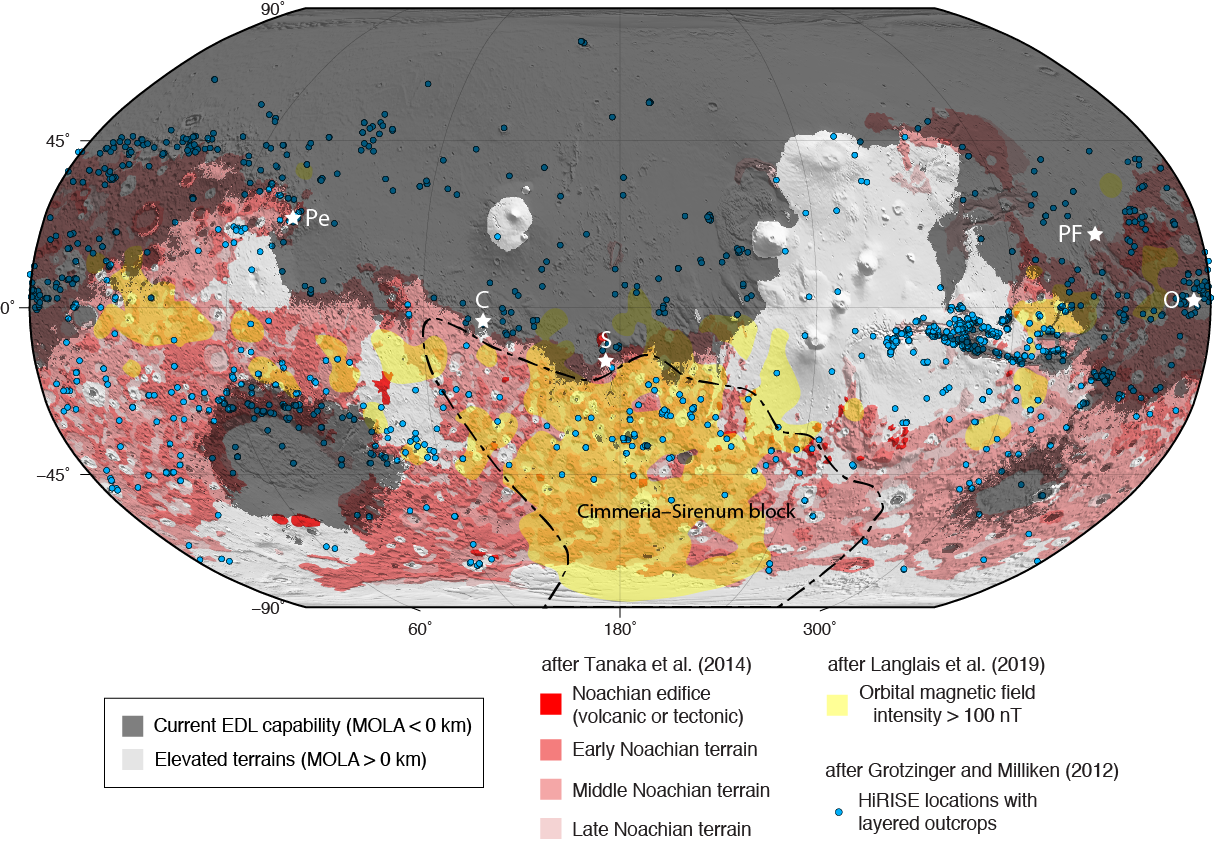}
\caption{Mid-air deployed (MAD) helicopters can land above current EDL technologies, at MOLA > 0 km, a decisive advantage to investigate the record of: paleomagnetism (in yellow area of crustal magnetic field >100 nT based on orbital measurements by MGS and MAVEN \cite{langlais_new_2019}), crust formation (Noachian aged-terrains in red \cite{tanaka_geologic_2014} and outline of Cimmeria-Sirenum block \cite{bouley_thick_2020}) and paleoenvironments (dots correspond to stratified deposits observed in HiRISE data \cite{grotzinger_sedimentary_2012}).}
\label{fig:map}
\end{figure}

The highlands terrains cover most of the southern hemisphere of Mars and present features relevant to each of the high priority science questions (Fig.~\ref{fig:map}):

Science priority 1: Paleomagnetism. Establishing the timing, duration, and strength of the Martian dynamo magnetic field is critical to understanding a variety of important geologic processes—from the thermal evolution of Mars to the escape of Mars’ atmosphere and long-term habitability. Strong crustal fields are mostly found in highland terrains, such as on the Terra Sirenum block \cite{bouley_thick_2020} (Fig.~\ref{fig:map}). This type of ancient crustal structures may be the result of early crustal differentiation mechanisms and tectonic history. There is currently no data on the remanent magnetic field that would help constrain the evolution of the dynamo and correlate it with processes of crust formation and climatic evolution recorded in sedimentary sequences. Acquiring the first \emph{in situ} stratigraphic analysis of the remanent magnetic field on key Noachian outcrops would yield a significant impact on our understanding of early Mars evolution.

Science priority 2: Magmatism/tectonism. The degree of crustal differentiation and the question of early crustal recycling on Mars has not been settled, yet this topic is fundamental for understanding the early Mars system because magmas drive the flux of volatiles from the interior, thus being a major contributor to climatic evolution. \emph{In situ} data on the lithology of key Noachian crustal components of the highlands (Fig.~\ref{fig:map}) are now needed to test models for: (i) the origin of the global crustal dichotomy, (ii) major processes forming the early crust. With the growing evidence for feldspathic and silica-rich magmatic rocks in the southern hemisphere \cite{carter_ancient_2013,wray_prolonged_2013,sautter_situ_2015}, it is possible that a significant fraction of the crust forming the southern highlands has a more evolved composition resembling the magmatic series of the Archean proto-continental crust. In addition, large scale geomorphic evidence suggests that a form of early crustal recycling could have occurred, such as the orogeny associated with the Thaumasia plateau \cite{anguita_evidences_2006,dohm_claritas_2009}. Such models add to the collective lines of evidence towards major shifts in our understanding of early Mars history and need to be tested \emph{in situ}.

Science priority 3: Paleoenvironments. Sedimentary sequences, often several kilometers thick, are found in terrains of both Noachian and Hesperian age, all around Mars’ surface (Fig.~\ref{fig:map}). Yet, the vast majority of these sequences remain of unclear origin. A global inventory of stratified outcrops from orbit highlighted six large "global" groups \cite{grotzinger_sedimentary_2012}. This grouping suggests that a small number of \emph{in situ} missions to key ancient sedimentary strata can advance our knowledge on early paleoenvironments with a global significance. Reference sedimentary sequences of Noachian age can be found at numerous locations on the highlands. Together, larger \emph{in situ} stratigraphic datasets on paleoenvironments will help understand how the nature and habitability of Mars persisted or evolved in conjunction with other early global processes such as volcanic degassing and loss of Mars’ atmosphere to space.

Helicopters are particularly suited for addressing these priorities given their ability to traverse rougher terrains than rovers. Furthermore, the capability of rotorcraft to cover 100+ km over a mission lifetime is crucial to document the diversity of features available on ancient surfaces, and to balance against the scarcity of naturally occurring high quality outcrop exposures.

Finally, although not unique to the Martian highlands, the regional-scale \emph{in situ} exploration capabilities of the helicopter could also close knowledge gaps related to regional atmospheric processes and modern day water cycling/soil formation (Science priority 4).

Table \ref{tab:inst} shows preliminary payload options, including 540-g (package A) that enable threshold observations for all top science priority objectives. Additional instruments would enhance science return (package B with ~2 kg). Payload mass will have to be traded against single flight range (Fig.~\ref{fig:range-vs-hover}). Mobility is a key capability of helicopter exploration, therefore, payload mass remains flexible within defined ranges as the MAD rotorcraft architecture will evolve towards most optimal design options.

\begin{table}[hbt!]
    \caption{Preliminary science payload options, for two possible packages: A at 540 g, and B at 2 kg}
    \centering
    \begin{tabular}{lcclc}
        \hline
        Instrument & Science & Mass (g) & Heritage & Package \\
        \hline
        Magnetometer & 1 & 20 & ESA cubesat Imperial College, UK & A,B \\
        Near infrared camera & 2 & 120 & Based on commercial InGaAs cameras & A,B \\
        AOTF spectrometer & 2,3 & 400 & Flying on Mars 2020 SuperCam & A,B \\
        Deployed XRF + Mossbauer & 2,3 & 1300 & Similar to MER & B \\
        Environment sensors & 4 & 100 & Similar to MSL REMS & B \\
        Soil sensors & 4 & 100 & Similar to Phoenix TECP & B \\
    \end{tabular}
    \label{tab:inst}
\end{table}

\section{Helicopter Design}
\label{sec:heli}

For the purpose of this study, we designed a helicopter with the capability of carrying a 540-g payload at a 0.01~kg/m$^3$ reference
atmospheric density. The high altitude optimizations not only allow for flight in the highlands, they also provide enough lift to
decelerate the helicopter after the MAD maneuver. We will refer to this helicopter as the MAD helicopter in this paper.

\begin{figure}[hbt!]
\centering
\includegraphics[width=.5\textwidth]{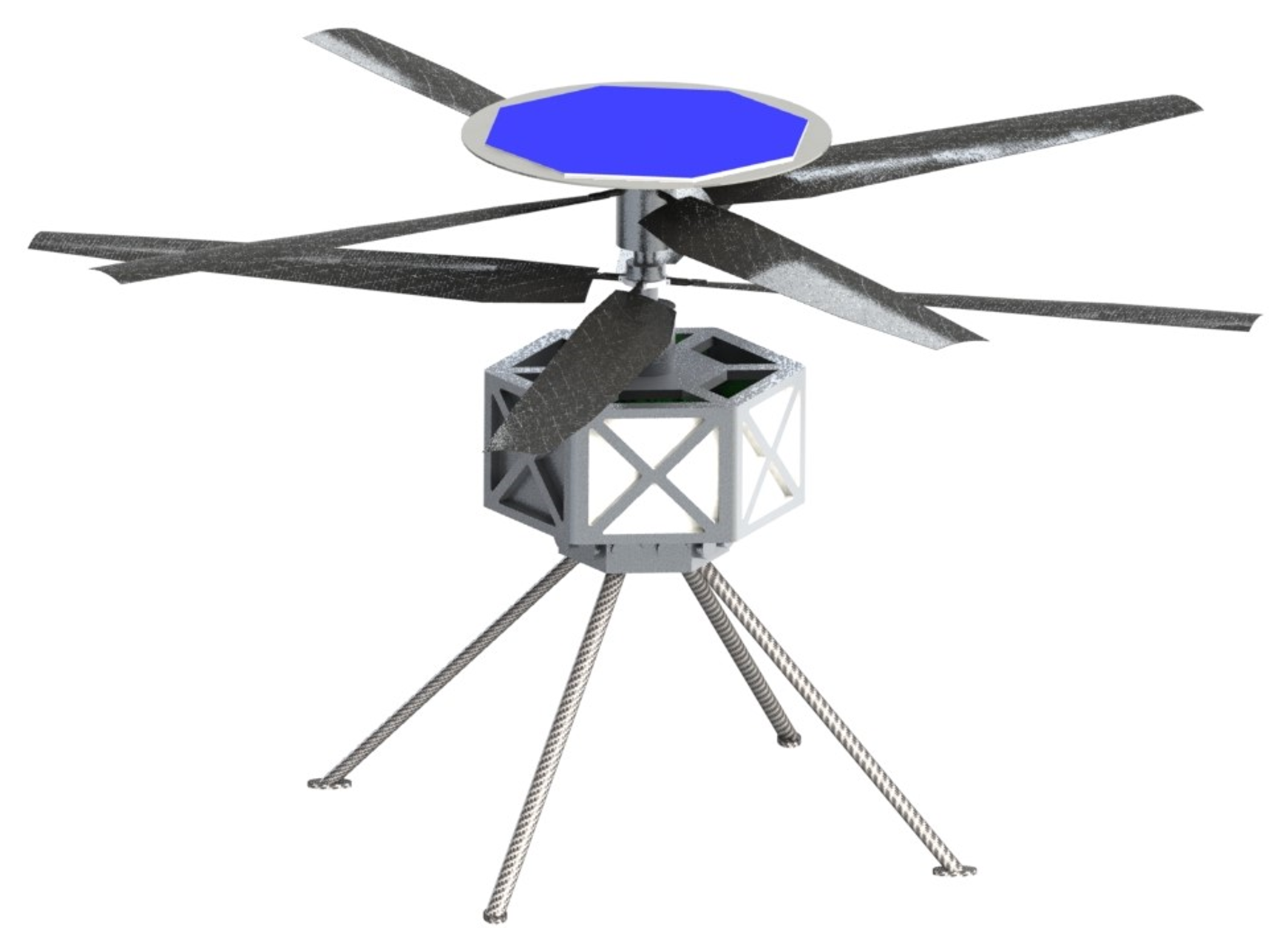}
\caption{Advanced Mars Helicopter design capable of carrying a 1.3-kg science payload at similar elevation than Ingenuity. The MAD
        helicopter design is similar with a larger blade area, and is capable of carrying a 0.54-kg science payload in the highlands.}
\label{fig:adv-mh}
\end{figure}

An on-going effort between JPL and Ames is focused on the design of Mars rotorcraft capable of carrying a science payload, at the similar 0.015 kg/m$^3$ density
Ingenuity will flying in in the Jezero crater~\cite{Withrow2020,Johnson2020}. One of these designs, which we will refer to as the \emph{Advanced Mars Helicopter} (MH),
is capable of carrying a 1.3-kg payload. It leverages Ingenuity flight heritage since it has the same size and configuration. Table~\ref{tab:helis}
compares the design parameters of Ingenuity, the Advanced MH and the MAD helicopter. One can see that the Advanced MH improves the payload capability, flight
range and hover time over Ingenuity by increasing the number of blades from two to four, the ratio of thrust coefficient to solidity $C_T/\sigma$ from 0.1 to 0.115,
the rotor solidity $\sigma$ from 0.148 to 0.248, and blade tip Mach velocity $M_{tip}$ from 0.7 to 0.8. The Advanced MH design is illustrated in Fig.~\ref{fig:adv-mh}.
\begin{table}
    \caption{Comparison between the \emph{Ingenuity} technology demonstration helicopter, an advanced Mars Helicopter (MH) designed to carry
        a science payload at similar atmospheric density, and the MAD helicopter designed to carry a science in the lower density encountered at high elevation.
        $C_T/\sigma$ is for the 2-rotor coaxial system.}
    \centering
    \begin{tabular}{llccc}
        \hline
        & & \emph{Ingenuity} & Advanced MH & MAD \\\hline
        Design $C_T/\sigma$  & & 0.1 & 0.115 & 0.095 \\
        Maximum $C_T/\sigma$ & & 0.135 & 0.135 & 0.161 \\
        Design $M_{tip}$ & & 0.7 & 0.8 & 0.8 \\
        MAD $M_{tip}$ & & N/A & N/A & 0.85 \\
        Design density & [$kg.m^{-3}$] & 0.015 & 0.015 & 0.01 \\
        Design temperature & [$C$] & -50 & -50 & -59 \\
        Cruise speed & [$m.s^{-1}$] & 2 & 30 & 30\\
        Payload & [$kg$] & 0.0 & 1.3 & 0.54 \\
        Range OR & [$km$] & 0.18 & 2.85 & 6.4\\
        Hover time & [$min$] & 1.5 & 6.6 & 2.85 \\
        R & [$m$] & 0.605 & 0.605 & 0.605\\
        Number of rotors & & 2 & 2 & 2 \\
        Number of blades & & 2 & 4 & 4 \\
        Gross weight & [$kg$] & 1.8 & 4.6 & 4.141 \\
        Disk loading & [$kg.m^{-2}$] & 0.8 & 2.0 & 3.6\\
        Solidity $\sigma$ & & 0.148 & 0.248 & 0.404 \\
        Rotor speed & [$rpm$] & 2575 & 2943 & 2882\\
        Total battery capacity & [$Wh$] & 44.4 & 170.2 & 119.1 \\
        \hline
    \end{tabular}
    \label{tab:helis}
\end{table}

\begin{figure}[hbt!]
\centering
\includegraphics[width=.5\textwidth]{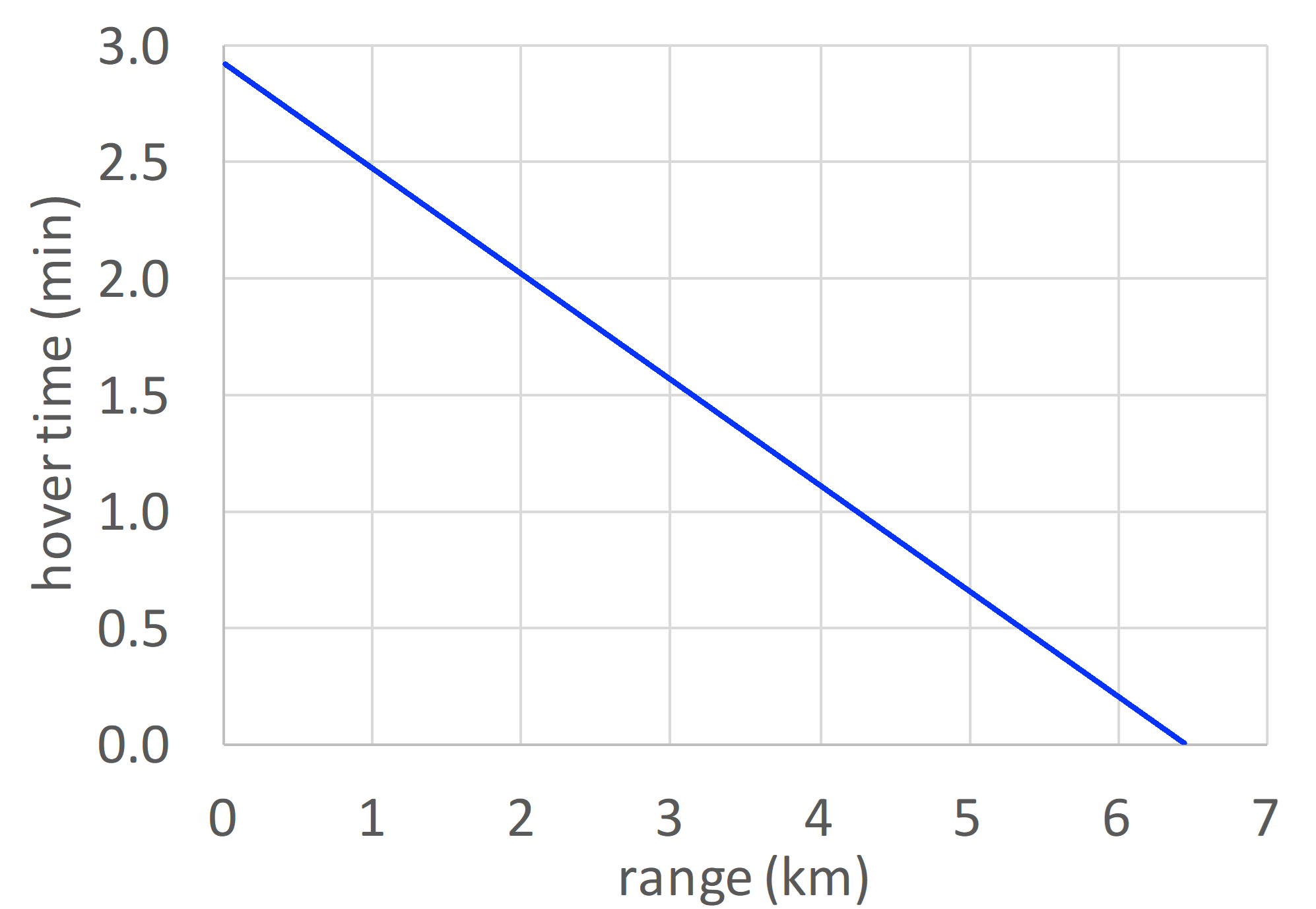}
\caption{Hover time versus range performance for the MAD helicopter.}
\label{fig:range-vs-hover}
\end{figure}

Similarly, the MAD helicopter improves upon the Advanced MH to optimize flight at high altitude by increasing $C_T/\sigma$ from 0.115 up to 0.161, and
$M_{tip}$ from 0.8 to 0.85 during MAD. It also increases the rotor solidity from 0.248 to 0.404. The range and hover performance of the MAD helicopter
are illustrated in Fig.~\ref{fig:range-vs-hover}.

Our companion paper~\cite{Young2020} analyzes the aerodynamics of the MAD helicopter and variants of it in nominal flight, as well as in proximity to the ground,
using Computational Fluid Dynamics (CFD) simulations. 

Now that we have established the weight and the dimensions of the MAD helicopter, the next section will focus on the entry and descent of this helicopter
on Mars to determine the velocity and altitude at which MAD needs to happen.

\section{Entry and Descent Analysis}
\label{sec:EDL}

\begin{figure}[hbt!]
\centering
\includegraphics[width=.5\textwidth]{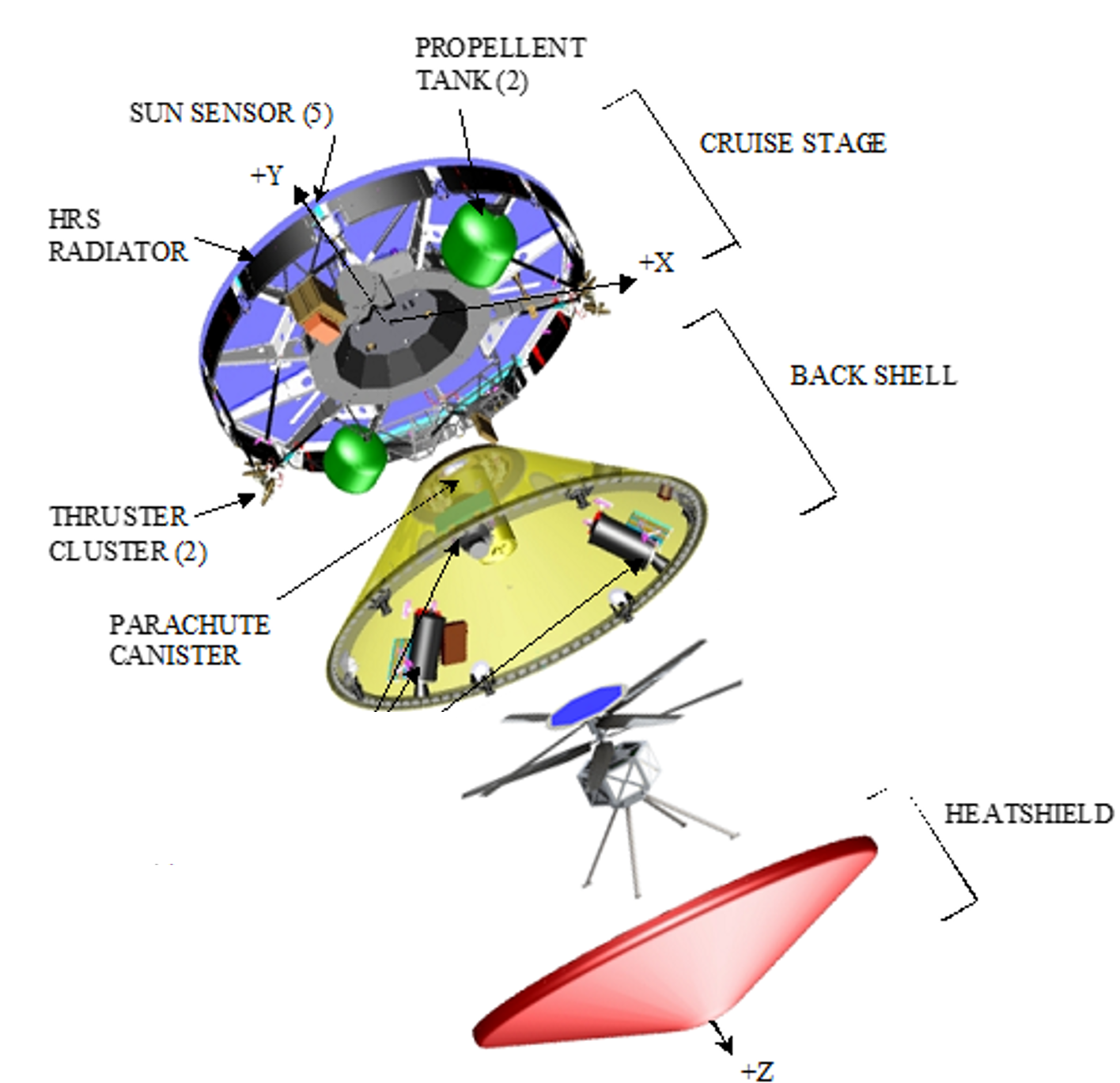}
\caption{Breakdown of the MAD architecture. The interplanetary cruise stage, the backshell, and the heatshield are heritage of the
\emph{Mars Exploration Rovers}. The helicopter is mounted on the backshell through a mechanism (not shown here, see Fig.~\ref{fig:mechanism}).}
\label{fig:edl-architecture}
\end{figure}

\begin{figure}[hbt!]
\centering
\includegraphics[width=\textwidth]{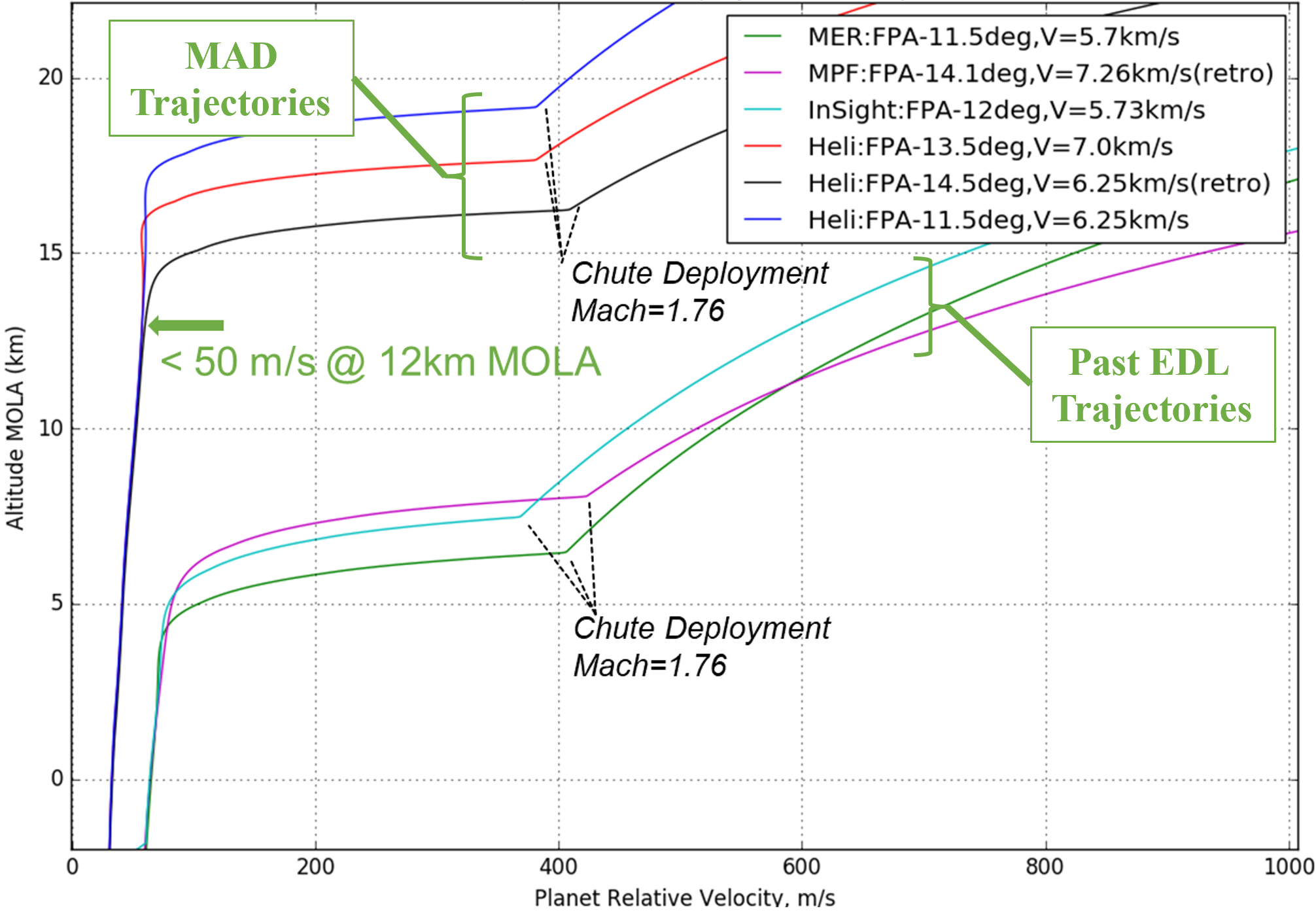}
\caption{Comparison of altitute versus planet-relative velocity between MAD and previous missions' EDL architectures. Using a similar heat shield geometry
    and parachute configuration, the mass reductions associated with the MAD architecture lead to sub-50 m/s velocities at 15 km MOLA, and a reduction of the
    terminal velocity to 30 m/s at Highlands elevations.}
\label{fig:edl-profiles}
\end{figure}

To maximize the probably of success, the MAD concept adopts the flight-proven elements of the existing EDL technology. Hence, in the proposed mission architecture for MAD adopts an aeroshell design heritage from Mars Exploration Rovers (MER), Mars Pathfinder (MPS) or Phoenix/InSight.
 All of the previous NASA Mars landers with the exception of MSL/M2020 and Viking have been designed to fit inside a 2.65-m aeroshell. From the flight dynamics perspective, a 70-deg spherecone heatshield has an extensive aerodatabase which can be utilized during the entry portion of the MAD EDL trajectory design. Based on the proposed concept of operations referenced in Fig.~\ref{fig:conops}, the terminal descent phase will be achieved via Disk-Gap-Band (DGB) supersonic parachute. The DGB parachutes have an extensive heritage for all of the Mars missions starting from the Viking I/II landers. The preliminary analysis has shown that the MER DGB parachute size with a nominal diameter of D0 = 14 m can provide a terminal velocity of $\sim$30 m/s for the MAD EDL configuration. Figure~\ref{fig:edl-architecture} illustrates the mechanical design of the MAD EDL system, with MER cruise stage and backshell components. Figure~\ref{fig:mechanism} provides some more details for the helicopter to backshell attachment points and deployment mechanism.  All of the MAD EDL design elements have strong flight heritage which can reduce both time and cost during the development cycle in Phases C, D of the mission.  Table~\ref{tab:edl} illustrates a direct mass comparison between the lightest flown Mars landers such as MPS, InSight and the MAD EDL concept. During the parachute phase of traditional Mars EDL, the lander travels at the terminal velocity of 63 m/s which requires to use a propulsive or an air-bag deployment phase to dissipate the required energy prior to a safe landing. The MAD concept, due to its light weight (i.e. entry mass of 256 kg) and an architectural decision of having a helicopter instead of a lander, eliminates the need of the propulsive phase. Furthermore, this architecture can significantly reduce the overall complexity of the EDL system. Figure~\ref{fig:edl-profiles} illustrates the EDL flight performance of the MAD EDL system compared to the MER/MPS and InSight missions. The EDL trajectories have been simulated with the Dynamics Simulator for Entry Descent and Surface Landing (DSENDS) JPL software. The Mars atmospheric conditions and winds were used identical for all of the illustrated trajectories in order to provide a direct comparison in altitude velocity space. The lightweight MAD configuration has shown a distinctive advantage, as the aeroshell can achieve the required DGB deployment conditions M < 2.0 at the altitude greater than 15 km MOLA. A smaller suspended mass (see Table~\ref{tab:edl}) allows to reach the terminal velocity of 30m/s significantly higher in the descent profile, which opens a new design mission design opportunity for a successful landing at higher MOLA elevations. The previously-flown small landers such as MPS/MER, due to the entry mass and parachute design constraints ,were forced to a substantially lower parachute deployment altitude and hence a maximum site elevation of -1.4km (MOLA)~\cite{Braun2006}. 
\begin{table}
    \caption{Comparison with MAD EDL scenario, the Mars Pathfinder mission, and the Mars Insight mission.}
    \centering
    \begin{tabular}{llccc}
        \hline
        EDL Mission Scenarios & & \emph{Pathfinder} & \emph{Insight} & \emph{MAD} \\\hline
        Max Entry Velocity & [Inertial, km/s] & 7.26 (retrograde) & 6.3 & 7.3 \\
        Entry Flight Path angle & [deg] & -14.1 & -12.0 & -14.5 $\rightarrow$ -11.0 \\
        Entry Mass & [kg] & 586.7 & 625.0 & 256.0 \\
        Heatshied mass & [kg] & 70.0 & 74.4 & 70.0 \\
        \{Backshell + Chute\} mass & [$kg$] & 145.0 & 115.6 & 145.0 \\
        Total landed mass & [$kg$] & 370.0 & 384 & 4.141 \\
        Terminal velocity on chute & [$kg$] & 63 & 63 & 30 \\
        \hline
    \end{tabular}
    \label{tab:edl}
\end{table}

The most critical phase in the MAD concept is the helicopter deployment in a freestream velocity of 30m/s. This is the focus of the next section.

\section{Deployment Design}
\label{sec:mad-design}

\subsection{Concept trades}

\begin{figure}[hbt!]
\centering
\includegraphics[width=\textwidth]{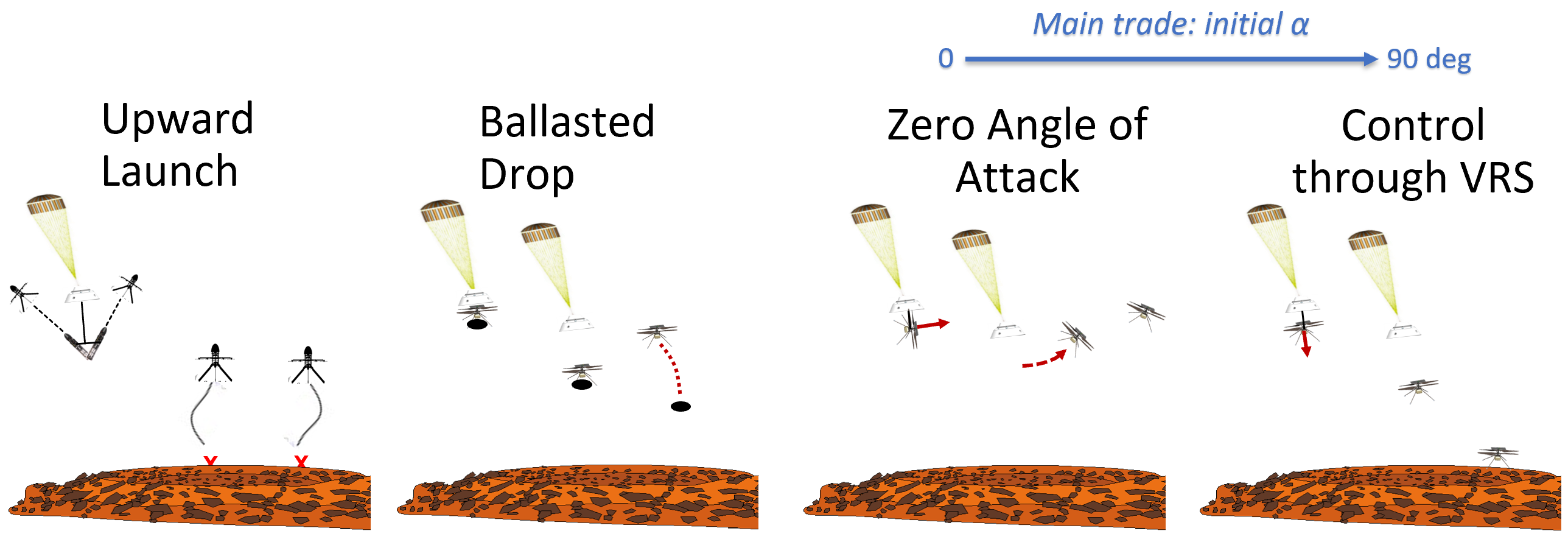}
\caption{Alternatives concepts considered for MAD: retrorocket assist (far left), helicopter attached to the heatshield (center left),
angle of attack $\alpha = 0$ deg (center right), $\alpha = 90$ deg (far right).}
\label{fig:concept-trades}
\end{figure}

Various options are available to deploy a helicopter stored in a backshell under a parachute at 30 m/s, see Fig.~\ref{fig:concept-trades}. Rocket propulsion
can provide the total delta-V to eliminate the terminal velocity before release. This solves the aerodynamics problem of MAD, but it also requires
3-axis attitude control of the backshell. A backshell architecture capable of 3-axis rocket-based attitude control will introduce complexity and cost
levels that are likely incompatible for an affordable Mars mission, e.g. within the NASA Discovery program. Another option is to have the helicopter
attached to the heatshield when it is jettisoned, so it can act as a ballast. A ballast increases the ballistic coefficient difference between the 
helicopter assembly and the backshell, which minimizes the risk of recontact. However, guaranteeing aerodynamic stability the heatshield is difficult,
the rotors would probably need to be spun up before separation so they are not as fragile, and this approach would require an externally-mounted backshell
altimeter to trigger MAD at the right height above the terrain. Furthermore, large scale flow separation downstream of the heatshield can be challenging for
the helicopter to maintain the desired attitude during separation

\begin{figure}[hbt!]
\centering
\includegraphics[width=.5\textwidth]{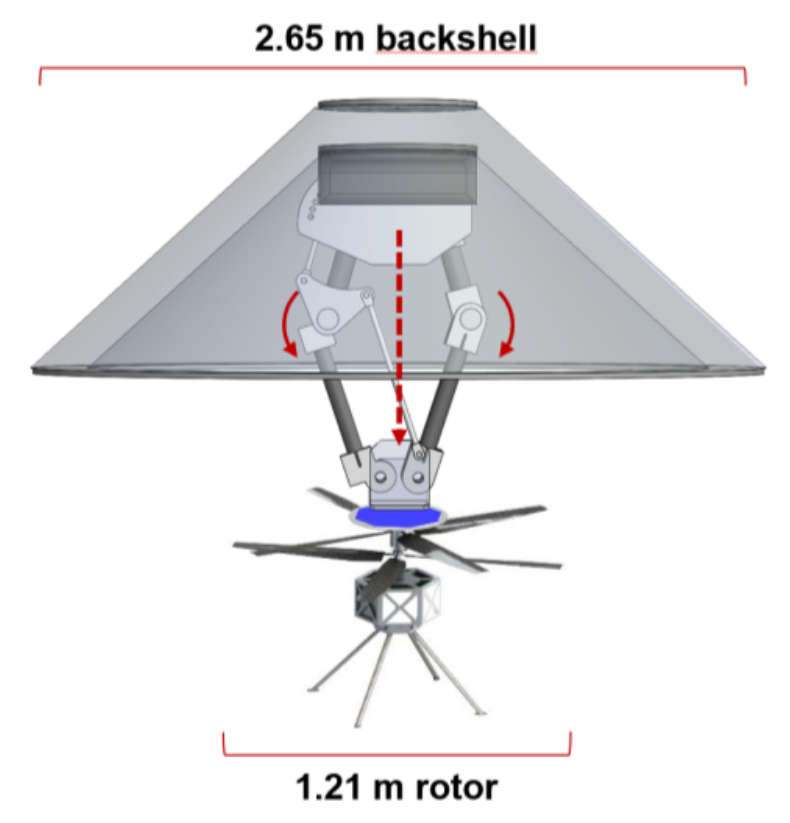}
\caption{Current backshell deployment mechanism baseline design.}
\label{fig:mechanism}
\end{figure}

Ballistic coefficients were estimated at 1.56 for the {backshell+chute}; and 3.24 for the MAD helicopter in autorotation. This high difference
provides high chances of successful separation. Rotors can be spun up at zero pitch while attached to the backshell, to provide
rotor controls as soon as the helicopter is released. Flow conditions within and in the vicinity of the open backshell are shown to be near-stagnant in
our companion paper~\cite{Young2020}, which supports the assumption that the rotorcraft can be controlled in these conditions. Figure~\ref{fig:mechanism}
illustrates our current baseline backshell design, featuring a linkage mechanism that has the ability of extending the helicopter past the lower skirt of the backshell with attachment points at the top of the rotorcraft. This not only allows the rotor to be spun up, the mechanism can provide an initial delta-V at separation too. Using
the collective blade pitch angle to provide negative lift at deployment is another possible source of delta-V~\cite{Young2020}. 

\subsection{Backshell deployment design}
\begin{figure}[hbt!]
\centering
\includegraphics[width=.75\textwidth]{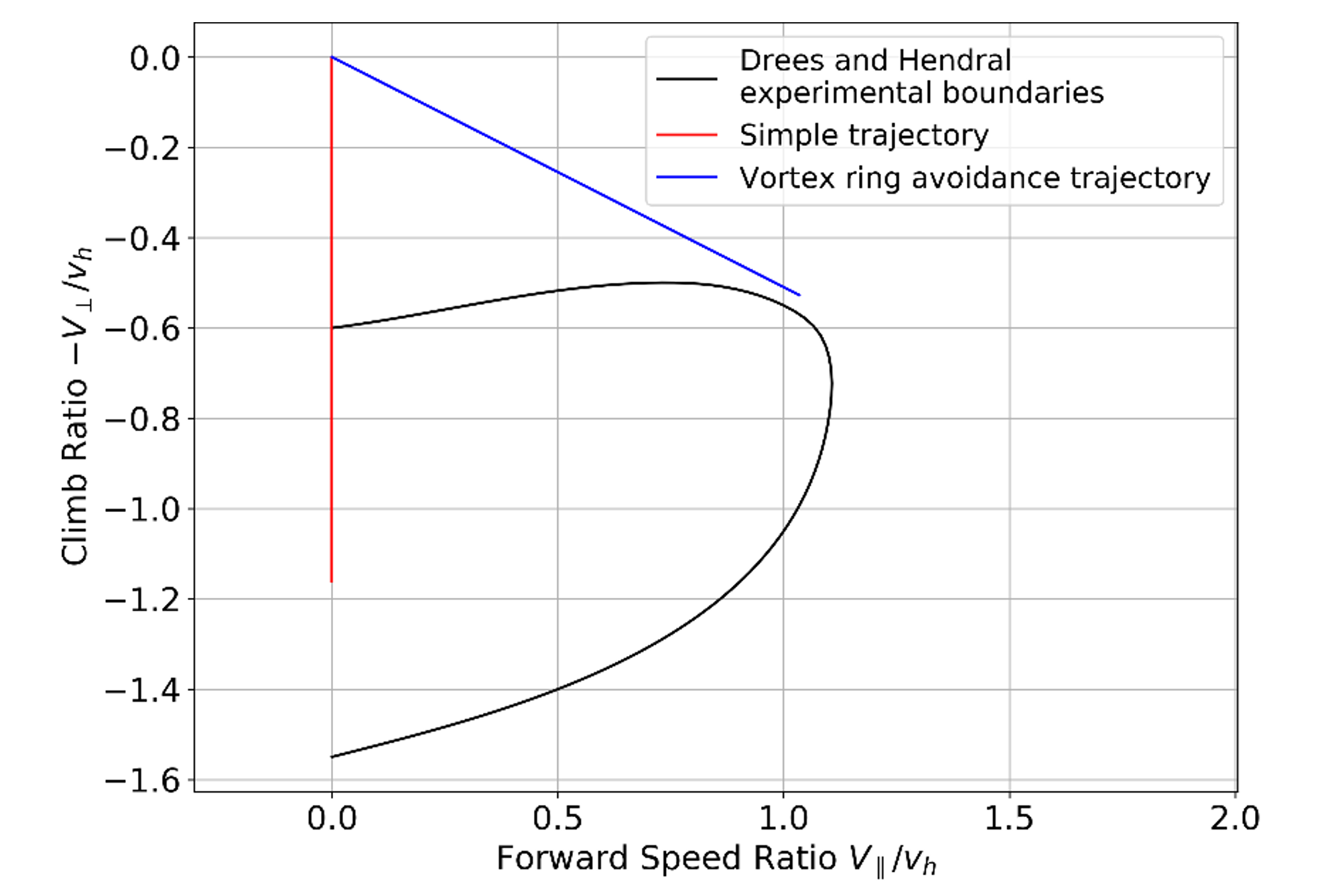}
\caption{Boundaries of the Vortex Ring State (VRS) regime in rotor-aligned velocity space, normalized the rotor induced
velocity at hover $v_h$. The constant $\alpha = 90$ deg angle of attack trajectory (red) provides maximum braking efficiency, but it has to face VRS conditions.
Safe trajectories (e.g. in blue) can be designed using lower values of $\alpha$ deg to avoid VRS, but they suffer increased altitude
loss.}
\label{fig:vrs-traj}
\end{figure}

Rotorcraft have complex aerodynamics during descent. Flow conditions such as the Vortex Ring State (VRS) and Turbulent
Wake State (TWS) can create variations of rotor thrust, which results in vibrations and velocity drops in single and coaxial rotorcraft
like the MAD helicopter~\cite{JohnsonVRS}. The boundaries of VRS/TWS are well-understood from experiments. The terminal velocity of the {parachute+backshell}
system can expose MAD to VRS/TWS based on the angle of attack of the rotor plane and the induced velocity at hover $v_h$ of the rotorcraft.
Figure~\ref{fig:vrs-traj} shows the normalized boundaries where VRS/TWS can happen. These are constant for all rotorcraft, since the velocity is normalized by $v_h$,
which is a function of the rotor geometry, thrust and air density. The red trajectory corresponds to upright configuration, where the helicopter is released
vertically and controlled at a angle of attack $\alpha = 90$ deg throughout the initial part of the trajectory. This
corresponds to the far right illustration in Fig.~\ref{fig:concept-trades}. This approach provides high-braking efficiency but stresses the controls system, which
has to be robust enough to handle VRS/TWS. Conversely, for a given deployment velocity magnitude, a lower $\alpha$ can be commanded to result in VRS-safe trajectory (blue in
Fig.~\ref{fig:vrs-traj}), at the cost of braking efficiency. This results in a guidance problem, where the desired solution minimizes the altitude loss to bring
the rotorcraft from MAD speed to hover.

The tradeoff between controlling the MAD helicopter through VRS/TWS, or commanding lower-$\alpha$ deg trajectories that alleviates
the aerodynamic turbulence is the main design driver of our approach. Our baseline mechanism, or variants using the attachment
point at the bottom of the rotor axis instead of the top, can enable the initial angle of attack at helicopter release to vary between 0 and 90 deg. The velocity
requirement on the descent trajectory ultimately depends on the controllability and stability of the helicopter in the descent flight
regime, extending the work in~\cite{Grip2018}. This will define exclusion zones within the VRS/TWS boundaries of Fig.~\ref{fig:vrs-traj},
around which an optimal trajectory can be designed~\cite{Talaeizadeh2020}. This is out of the scope of the current paper. The next section will
discuss preliminary simulation results with a simplified rotorcraft dynamics model and controls.

\section{Preliminary Evaluation of Mid-Air Deployment Performance}
\label{sec:mad-sim}

\subsection{Simulation framework}

The end-to-end MAD simulation was created using the general-purpose DARTS simulation toolkit~\cite{DARTS1}, which provides the framework 
for modeling and simulation of the entry, parachute, and rotorcraft phases. DARTS is designed to handle rigid/flexible
multibody dynamics, arbitrary system topologies, smooth and non-smooth dynamics, and
run-time configuration changes such as the MAD parachute deployment and rotorcraft release
events. The toolkit provides a full complement of computational algorithms for dynamics analysis and model-based
control with fast computational performance using low-cost recursive algorithms
for solving equations of motion. DARTS computational algorithms are structure-based and consist of scatter/gather recursions that proceed
across the bodies in the system topology. This allows DARTS to be a general-purpose tool requiring no
change to the software to model multibody systems with arbitrary numbers of bodies and branching structure. This property also allows DARTS to easily handle run-time structural changes in the system topology,
such as the attachment/detachment and addition/deletion of bodies. While the DARTS object-oriented implementation is in C++,
a rich Python interface is available for all classes and methods in the system. This allows
users full flexibility in defining and configuring the model as desired.

DARTS also provides a component model-based simulation framework to add and couple parameterized models 
for actuator/sensor devices and the
environment interactions needed for modeling the overall physics of a vehicle. The re-usability of component models is key to the usage of DARTS across multiple platform applications, and even
platform domains (e.g., aerospace, Unmanned Ground Vehicles (UGV), Unmanned Aerial Vehicles (UAV), robotics). The component model library within DARTS includes
sensors (e.g., Inertial Measurement Unit (IMU), cameras, Laser Imaging Detection and Ranging (LIDAR)), actuators (e.g., motors, engines), and environment interactions (e.g.,
gravity, aerodynamics, terramechanics) that can be used out of the box, adapted, or added to third
party sources as needed. For instance, the aerodynamics models rely on planetary atmospheric models using the  Global Reference Atmospheric Model (GRAM) and wind models to compute the aerodynamics forces on flight vehicles. 

\subsection{MAD models}

The helicopter simulation utilizes the DARTS multibody dynamics simulation framework described earlier to implement a coaxial helicopter multibody structure containing a base, and two rotors to be equivalent
to the MAD helicopter of Table~\ref{tab:helis}. In addition, models are integrated into the simulation paradigm that calculate and apply the lift, drag, and gravity forces acting on the multibody structure.

The MAD simulation takes three command inputs:
\begin{itemize}
    \item the coefficient of thrust to solidity ratio, $C_T/\sigma$, which in practice would be controlled using the collective blade pitch angle;
    \item the motor torque $Q_M$;
    \item an attitude control torque $Q_{att}$, which in practice would be controlled using cyclic blade pitch angle.
\end{itemize}
A full rotor model including, collective, cyclic and swash plate was out of the scope of this paper. The equations in this section assuming a single rotor
disk actuator, as assumed in momentum theory~\cite{JohnsonHelicopterTheory}, to model the effect of the two coaxial rotors. The simulation currently assumes
that the two rotor produce the same trust and have the same rotation rate. The values of $T$ and $Q$ are divided by $2$ before being applied to each rotor, and
the torque is applied in opposite direction for each rotor.

$C_T/\sigma$ is assumed to be ideally controlled to the
constant MAD design value specified in Table~\ref{tab:helis}. The rotor is assumed to be spinning at a rotation rate $\Omega$ from the start of the
deployment maneuver, so that blade tip speed readily corresponds to the MAD design value $M_{tip}$ from Table~\ref{tab:helis}. Given $C_T/\sigma$ and $\Omega$,
the overall thrust for the two-rotor system is computed using
\begin{align}
    T &= C_T \rho A (\Omega R)^2 
\end{align}
where $T$ is the thrust, $A$ is the rotor-disk area, and the remaining symbols are defined in, or in terms of, the values given in Table \ref{tab:heliParams}.

The aerodynamic torque on the rotor is calculated using
\begin{align}
    Q &= C_Q \rho A R (\Omega R)^2\\
    C_Q &= \frac{\overline{C_D}\sigma}{8}\bigg(1+\Big(\frac{6C_T}{\sigma}\Big)^2+\bigg(\frac{\frac{C_T}{\sigma}}{(\frac{C_T}{\sigma})_s}\bigg)^{n_s}\bigg)(1+4.6\mu^2)+C_T\lambda
    \label{eq:cq}\\
    \mu &= \frac{V_x}{\Omega R} \\
    \lambda &= \frac{V_z+v_i}{\Omega R}
\end{align}
where $Q$ is the torque on the rotor, $V_x$ is the velocity of the helicopter parallel to the rotor-disc, $V_z$ is the velocity of the helicopter perpendicular to the rotor-disc, and the remaining symbols are defined in, or in terms of, the values given in Table \ref{tab:heliParams}. The induced velocity $v_i$ is calculated using the VRS model from~\cite{JohnsonVRS}, which allows $v_i$ to be computed in the vertical descent velocity regimes where rotor momentum theory is singular~\cite{JohnsonHelicopterTheory}. The motor torque $Q_M$ is commanded to match the aerodynamic torque $Q$ through a PI controller, and keep the blade tip speed constant.

The fuselage drag is calculated via
\begin{equation}
    \mathbf{D} = -\frac{1}{2}C_D\rho A_{base} V^2 \hat{\mathbf{V}}
\end{equation}
where $\mathbf{D}$ is the drag on the helicopter, $\mathbf{V}$ is the velocity of the rotorcraft with respect to the incoming flow, $V$ is the magnitude of $\mathbf{V}$, $\hat{\mathbf{V}}$ is a unit vector in the direction of $\mathbf{V}$, and $A_{base}$ is the reference area. The gravity model is a simple point mass model.
\begin{table}
\centering
\caption{Helicopter Parameters used during simulation with VRS Model (see nomenclature)}
\label{tab:heliParams}
\begin{tabular}{ll} 
	\hline
	Parameter & Value \\\hline
	$\overline{C_D}$ & 0.03 \\
	$\Big(\frac{C_T}{\sigma}\Big)_s$ & 0.20 \\
	$n_s$ & 20 \\
	$f$  & 1.0 \\
	$k$ & 1.1 \\
	$\Omega_{max}$ & 302 rad/s \\
	$\tau_{max}$ & 4.41 N-m \\
	$\rho$ & Based on MarsGram 2010 atmospheric profile  \\
	$v_w$ & Based on MarsGram 2010 atmospheric profile  \\ 
	$\theta_{max}$ & $21^{\circ}$ \\
	$C_D$ & 0.8 \\
	$s$ & 0.14 meters \\
	\hline
\end{tabular}
\end{table}

\subsection{Simulation results}

\begin{figure}[hbt!]
\centering
\includegraphics[width=\textwidth]{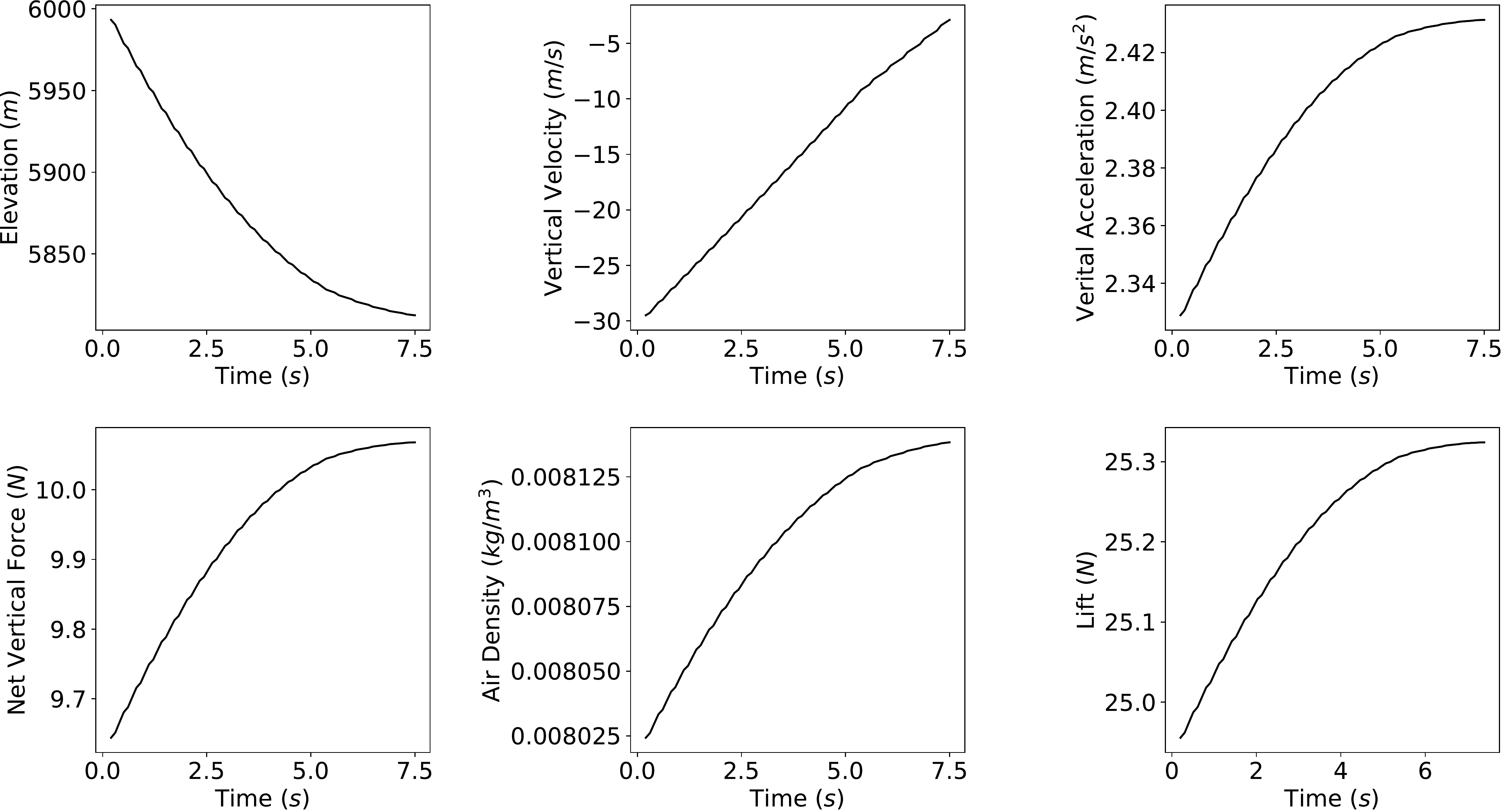}
\caption{MAD simulation results using the MAD helicopter design, DARTS simulation framework and a VRS model for rotor induced velocity.
From a start altitude of 6,000 m MOLA,the helicopter takes about 250 m to come to a hover, at a landing elevation of 5,750 m MOLA.}
\label{fig:sim-traj}
\end{figure}

Figure~\ref{fig:sim-traj} shows the MAD simulation results, assuming a velocity of 30 m/s at a release altitude of 6,000 m MOLA. This initial
altitude was chosen to be consistent with our preliminary simulation results using a point-mass helicopter model, with ideal controls and simplified
VRS dynamics and atmospheric model~\cite{Bhagwat2020}.
Density is provided by the nominal values of the 2010 Mars GRAM model~\cite{GRAM2010}. Throughout the entire flight, the helicopter attitude maintains a 90-deg angle of attack using a PD controller, similar to the red trajectory in Fig.~\ref{fig:vrs-traj}. It can be shown that the after an elevation loss of 250m the rotorcraft has undergone complete vertical deceleration and has come to a hover for a potential landing at an elevation of 5,750 m MOLA.

\begin{figure}[hbt!]
\centering
\includegraphics[width=.75\textwidth]{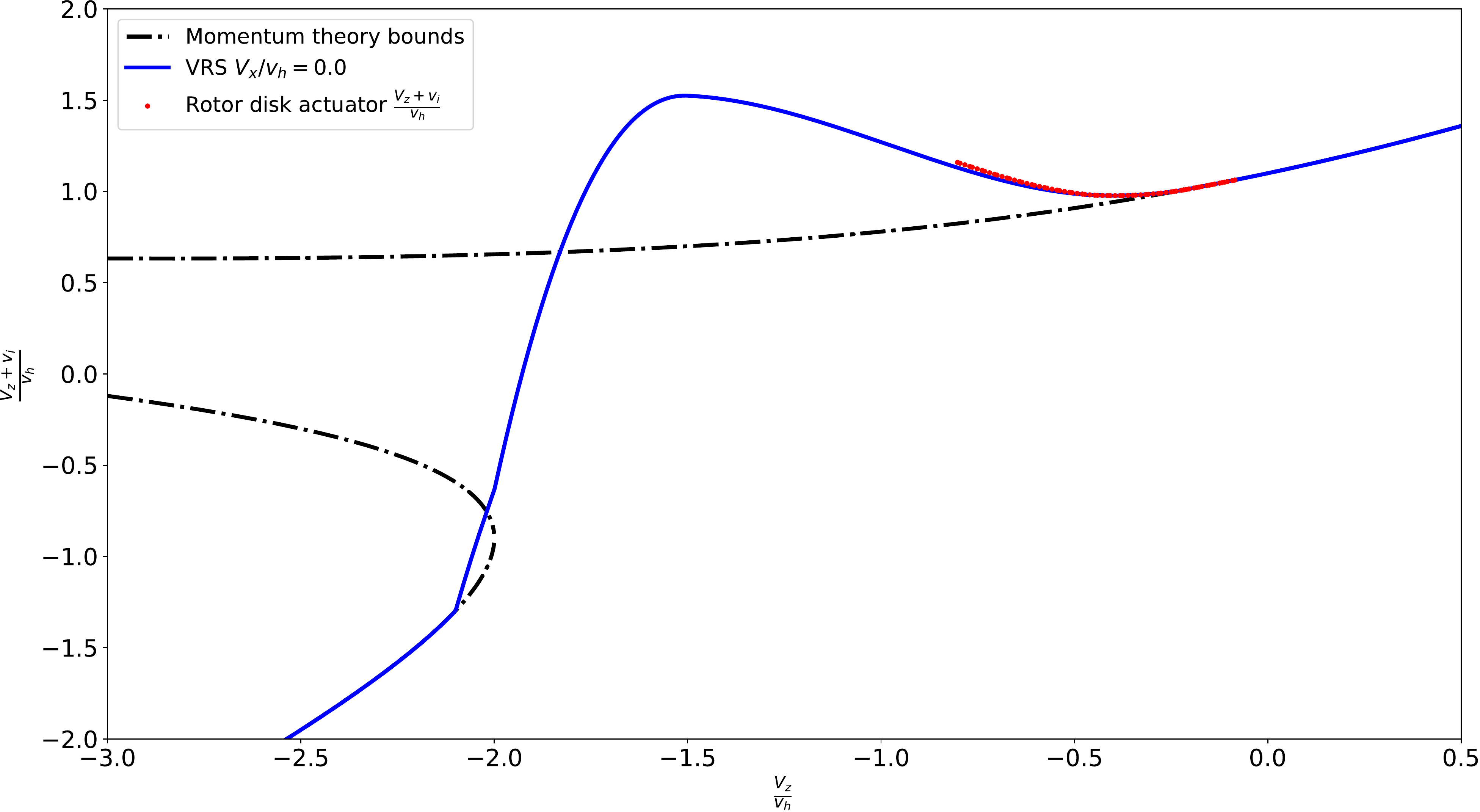}
\caption{Comparison between the VRS model for $\alpha = 90$ deg, and the simulated MAD trajectory. $v_i$ is the rotor induced velocity,
$V_z$ in the velocity component perpendicular to the rotor in the far field, $V_x$ is the velocity component in the rotor plane which
is null for $\alpha = 90$ deg, and $v_h = \sqrt{\frac{T}{2 \rho A}}$ is the ideal induced velocity at hover which is used as a normalization parameter.}
\label{fig:vrs-model}
\end{figure}

Figure~\ref{fig:vrs-model} shows the rotor induced velocity in the simulation, and overlays it on top of the momentum theory and VRS model
for $\alpha = 90$ deg deg. This plot confirms that the simulation is entering the VRS regimes, as the induced velocity progressively detaches from
the momentum theory curve. Since this simulation is assuming idealized instantaneous controls, VRS did not create vertical velocity oscillation
or drop-offs here, as it would in a real helicopter. It only manifests as a loss of efficiency as the motor needs to provide more torque to
compensate for the rising $v_i$ in Equation~(\ref{eq:cq}). A more advanced rotor model and realistic controls of the motor torque and
rotor collective is required to observe the trajectory effects of VRS, but is out of the scope of this paper.

Our companion paper~\cite{Young2020} further analyzes the aerodynamics of the MAD helicopter and variants of it in proximity to the backshell using CFD simulations.

\subsection{Preliminary Experimental Validation} \label{sub:experiments}
To gain experimental experience in MAD and provide additional supplementary data for the DARTS simulation, we have begun an experiment campaign in the Caltech Center for Autonomous Systems and Technologies (CAST) wind tunnel facility. The campaign consists of a consumer-off-the-shelf rotorcraft with variable pitch controls similar to the MAD helicopter design, released from a subscale backshell, and then flown under pilot control. The specific rotorcraft architecture was based on a quadrotor configuration, to facilitate operation in the confined lab environment. The facility features a vertical low speed wind tunnel system, which enables the simulation of arbitrarily long descents under controlled environments. It is also instrumented with a motion capture system for in-flight trajectory tracking.

As is often the case for scale testing, not all non-dimensional parameters can be matched simultaneously. Therefore, the most pertinent parameters for the investigated flow physics are prioritized. The most pertinent physics for this research are regarding the effect of VRS upon the rotor aerodynamics, which causes a change in control authority and thrust efficiency as helicopter encounters its own wake. The effect is primarily parameterized by the descent ratio $V_z/v_h$ which, as shown in Fig.~\ref{fig:vrs-model}, is predicted to vary from $0 < V_z/v_h < 1$ during the MAD descent. Accordingly, a variable-pitch rotorcraft was selected (rather than variable-RPM in order to keep rotor behavior consistent), and sized such that the range of achievable wind tunnel velocities $V_z$ and hover downwash $v_h$ fit in the specified descent rate range $V_z/v_h$.

\begin{figure}[hbt!]
\centering
\includegraphics[width=.75\textwidth]{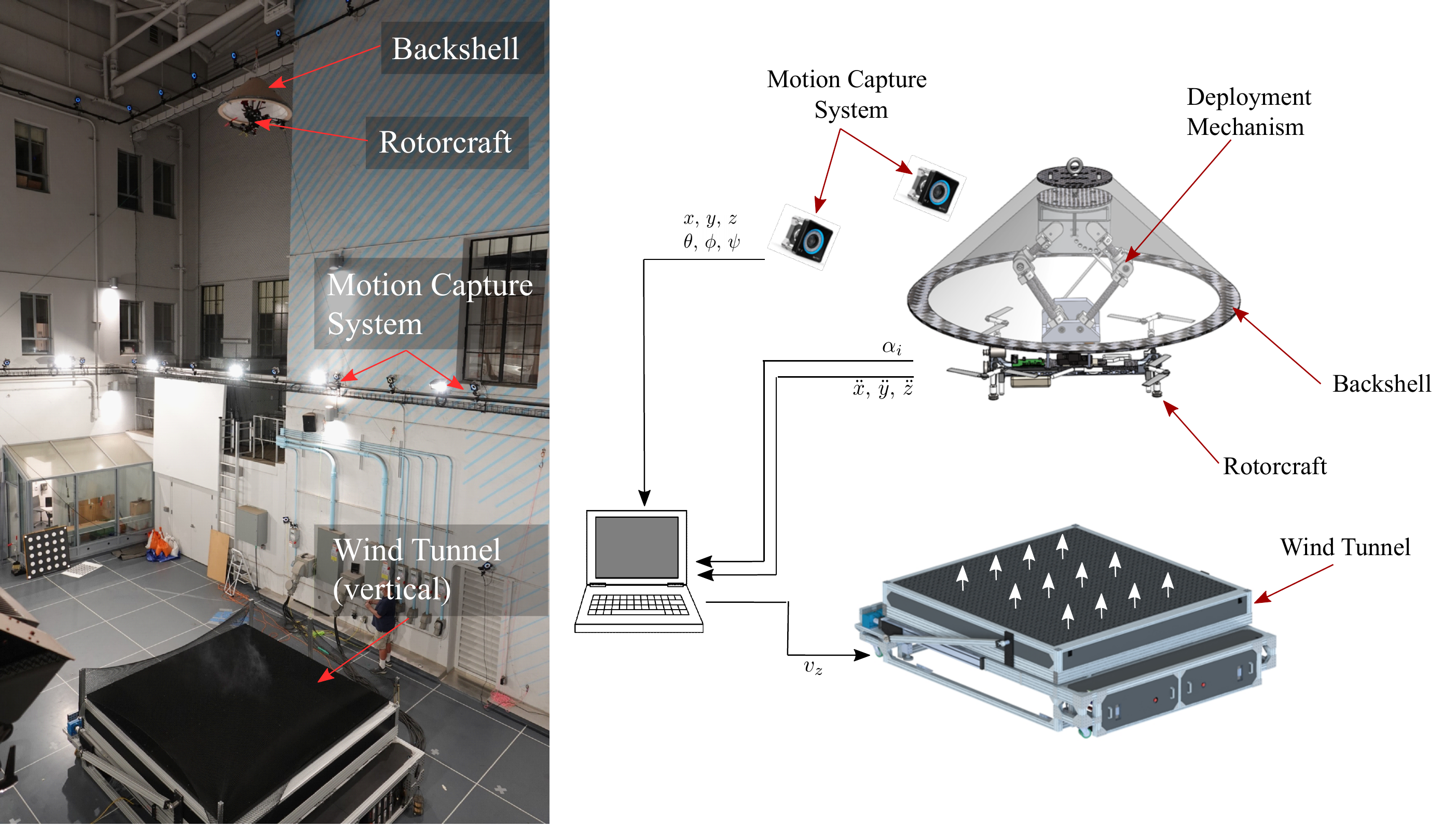}
\caption{Earth-Analog Experiment. (Left) Photograph of setup, including rotorcraft suspended above the vertical wind tunnel. (Right) Cartoon of setup, including data streams from the motion tracking system, the rotorcraft accelerometer, and the rotorcraft control actions (rotor collectives). Wind tunnel velocity is commanded from the control station, and rotorcraft controlled via radio link.}
\label{fig:experimental_setup}
\end{figure}

The experimental setup is illustrated in Fig.~\ref{fig:experimental_setup}. A 1:3 scaled Pathfinder backshell, constructed out of foam, is suspended 8 m above the wind tunnel by tether lines. A deployment mechanism, designed and fabricated into the volume-constraints of the backshell, is actuated by servomotors to extract the rotorcraft below the backshell, followed by radio controlled pin-puller mechanisms that release the helicopter into the wind tunnel flow. A centralized control station logs the rotorcraft pose via the offboard motion tracking system (obtaining the six-degree-of-freedom attitude of the rotorcraft), as well as data streamed from the rotorcraft (accelerometer, gyro, and rotor control actions). Instantaneous thrust measurements based on accelerometer data can be referenced to RPM and blade pitch measurements to quantify the loss of propulsive efficiency during descent. The drag of the fuselage, estimated from the known vehicle projected area, is being subtracted out from the acceleration measurements to isolate the rotorborne thrust.

To date, the setup has been constructed, calibration performed, and initial descending flights after release from the capsule have been flown. During initial flights of the rotorcraft hovering within the vertical freestream of the wind tunnel, we could verify a significant loss of propulsive efficiency of the vehicle, which is consistent with the velocity zone predicted by VRS models \cite{JohnsonVRS}, but fully autonomous deployments from the backshell into the wind tunnel freestream flow and comparisons with the simulated trajectories are considered future work.

\section{Conclusion}

This paper proposed a EDL concept for a Mars helicopter science mission, based on mid-air deployment of the helicopter from the backshell after the heatshield is jettisoned.
Our design has a strong flight heritage from the Ingenuity helicopter design, and uses the same heatshield and chutes as in standard Mars EDL. MAD could enable a breakthrough cost and complexity
reduction for future Mars rotorcraft missions. This could provide a significant competitive advantage for a future Discovery solicitation, or if new classes of Mars
missions are introduced between SIMPLEx and Discovery. MAD could enable an entirely new mission to regions previously inaccessible, like the Highlands; or supports existing
science helicopter mission concepts to lower elevations. The main challenge of our MAD approach is to safely spin up the rotor while it is still attached to the backshell, and
safely control the descent trajectory to maneuver through potentially turbulent descent conditions. We presented a science analysis that derived the minimum payload required to perform science in 
the Highlands, a helicopter that can fly that payload at these elevations, a mechanism that allows to safely spin up and clear the backshell, as well as preliminary end to
end simulation results from entry to touchdown. Preliminary results are consistent with the possibility to land above 5,000 m MOLA elevation.

Future theoretical and experimental work is required to deliver a proof of feasibility of MAD in Martian conditions. In simulation, the rotor model needs to account for the swashplate and blade dynamics.
Realistic controls need to be implemented for the motor torque, and blade pitch angle. And navigation with realistic sensor noise should be done in the loop to provide a representative evaluation
of the closed-loop performance. Performance trades need to be evaluated with this new simulation framework, to extend our previous work on the system trades~\cite{Bhagwat2020}. Once the concept is matured, more realistic CFD simulations should be run to confirm the results of our companion paper~\cite{Young2020}. Finally, wind tunnel
experiments can provide validation of the models we use in simulation. On-going experiments at Caltech/CAST have been focused demonstrating the release mechanism, and collecting VRS data over the
wind tunnels where the aerodynamics have been scaled to match the Mars case.

\section*{Acknowledgments}
The research described in this paper was carried out at the Jet Propulsion Laboratory, California Institute of Technology, under a contract with the National Aeronautics and Space Administration (80NM0018D0004).

This work was supported by JPL's Spontaneous R\&TD program, and a NASA Space Technology Research Fellowship, Leake [NSTRF 2019] Grant \#: 80NSSC19K1152.

The authors would like to thank Chad Edwards and Larry Matthies, from JPL's Mars Exploration Program Advanced Concepts Office, for their technical and programmatic support; as well as Theodore Tzanetos, Mars Helicopter tactical lead at JPL, for sanity checking Ingenuity's design parameters.

\textcopyright~2020. All rights reserved.

\bibliography{main}

\end{document}